\documentclass[conference]{IEEEtran}
\IEEEoverridecommandlockouts
\usepackage{cite}
\usepackage{amsmath,amssymb,amsfonts}
\usepackage{algorithmic}
\usepackage{graphicx}
\usepackage{textcomp}
\usepackage{pifont}
\usepackage{tabularx}
\usepackage{booktabs} 
\usepackage{adjustbox}
\usepackage{multirow}
\usepackage{xcolor}
\usepackage{subcaption}
\usepackage{url}
\def\BibTeX{{\rm B\kern-.05em{\sc i\kern-.025em b}\kern-.08em
    T\kern-.1667em\lower.7ex\hbox{E}\kern-.125emX}}
    
\begin{document}

\title{All-Optical Segmentation via Diffractive Neural Networks for Autonomous Driving}

%\author{Yingjie Li}
\author{\IEEEauthorblockN{Yingjie Li$^{*+}$}
\IEEEauthorblockA{\textit{} 
 \textit{Simon Fraser University}\\
yingjie\_li@sfu.ca} \\
\thanks{+: Corresponding author: Yingjie Li (\texttt{yingjie\_li@sfu.ca}).}
\thanks{*: Work done when studay at the University of Utah.}
 \and
 \IEEEauthorblockN{Daniel Robinson$^{*}$}
 \IEEEauthorblockA{\textit{}
  \textit{Massachusetts Institute of Technology}\\
 daniel\_r@mit.edu}
 \and
 \IEEEauthorblockN{Weilu Gao}
 \IEEEauthorblockA{\textit{}
  \textit{University of Utah}\\
 weilu.gao@utah.edu}
 \and
 \IEEEauthorblockN{Cunxi Yu}
 \IEEEauthorblockA{\textit{} 
 \textit{University of Maryland, College Park} \\
 cunxiyu@umd.edu}
 }

\maketitle

\begin{abstract}

Semantic segmentation and lane detection are crucial tasks in autonomous driving systems. Conventional approaches predominantly rely on deep neural networks (DNNs), which incur high energy costs due to extensive analog-to-digital conversions and large-scale image computations required for low-latency, real-time responses. Diffractive optical neural networks (DONNs) have shown promising advantages over conventional DNNs on digital or optoelectronic computing platforms in energy efficiency. By performing all-optical image processing via light diffraction at the speed of light, DONNs save computation energy costs while reducing the overhead associated with analog-to-digital conversions by all-optical encoding and computing.
In this work, we propose a novel all-optical computing framework for RGB image segmentation and lane detection in autonomous driving applications. Our experimental results demonstrate the effectiveness of the DONN system for image segmentation on the CityScapes dataset. Additionally, we conduct case studies on lane detection using a customized indoor track dataset and simulated driving scenarios in CARLA, where we further evaluate the model's generalizability under diverse environmental conditions.

\end{abstract}


%%%%%%%%%%%%%%%%%%%%%%%%%%%%%%%%%%%%%%%%%%%%

\section{Introduction}
\label{sec:introduction}

Autonomous driving represents one of the most transformative innovations in modern transportation. Autonomous driving is designed to perceive the environment, make decisions, and navigate roads with minimal human intervention. The development of autonomous driving systems not only benefits people's daily life but also shows the potential to enhance the road safety and improve the traffic efficiency. Autonomous driving has become one of the most actively researched and rapidly evolving fields with widespread attention from both academia \cite{mora2020mind,hu2023planning} and industry \cite{jiang2025scenediffuser,noomwongs2020design}. 

Within an autonomous driving system, the environment is first sensed by delicate sensors such as cameras, LiDAR, radar, etc. Then the sensor data is combined to create an accurate environment model. The environment is then analyzed by perception models, which are typically deep-learning based models, to build a complete scene understanding for the vehicle. After predictions and planning with the analyzed information from perception models, controls are generated to physically move the vehicle along the planned path\cite{yurtsever2020survey}.  
During the process, the sensed information is converted to digital format and fed to digital systems for processing. The design of the perception model with the sensor data is critical in autonomous driving. Many deep learning frameworks have been proposed as the perception models for reliable and energy-efficient perception in all environments and conditions, providing efficient and real-time response (autonomous control generation) to the environment\cite{wang2024yolov9,kirillov2023segment,wang2024rt}. 

However, the communication and memory access with digital processors are energy consuming, and can cause significant latency \cite{liu2022understanding}. Moreover, digital DNN frameworks require substantial computational efforts and expensive memory usage to achieve high performance, posing further challenges for edge computing and power management during the deployment of autonomous vehicles\cite{muhammad2020deep}. 

To enhance the efficiency of autonomous driving systems, it is important to improve the latency and the computational efficiency of large-scale digital DNNs.  
There has been a growing trend towards developing novel high-performance while energy-efficient DNNs platforms, especially implementing novel DNNs in the optical domain, e.g., optical neural networks (ONNs), that mimic conventional feed-forward neural network functions using light propagation \cite{lin2018all,shen2017deep,hu2024diffractive}.
{Free-space Diffractive Optical Neural Networks (DONNs) is a promising research area in ONNs for autonomous driving in that: \textbf{(1)} Different from digital computing platforms and integrated photonics-based optoelectronic systems that demand preprocessed inputs, free-space DONN systems have direct access to all the optical degrees of freedom that carry information about an input scene/object without needing digital recovery or preprocessing of information. \textbf{(2)} DONNs feature with its high system throughput, inherent parallel processing and fast computation speeds with minimal power consumption by enabling all-optical inference through passive optical devices. DONNs realize all-optical computation by manipulating the information carrier, light signal, with physical phenomena such as light diffraction and phase modulation, which occur by nature at the speed of light in the medium, while requiring no additional energy for computation.

%The all-optical free-space DONN system is a strong candidate for efficiently handling images-related machine learning (ML) tasks in that: \textbf{(1)} In DONN systems, the information is directly encoded onto the light signal, saving the need for ADC. While in digital DNN systems, it involves intense ADC and memory footprint for successive digital processing for the analog information captured by sensors, which can be the bottleneck in both inference speed and overall performance. 
%\textbf{(2)} DONN systems inherently enable efficient parallel computation across multiple channels at minimal energy cost through passive optical devices. For example, in conventional digital DNNs, processing RGB images requires triple energy and memory resources, compared to gray-scale images. While in DONN systems, passive optical components such as beam splitters can expand the light signal into separate channels without additional energy consumption.
%Although triple phase masks are still required in the system, their presence incurs only fabrication costs without ongoing energy expenditure. Once fabricated, the passive optical devices incur no extra energy cost for its functionality and the computation happens by nature in each channel at light speed. 

To bring the advantages of DONN systems into autonomous driving applications, we propose a novel DONN architecture designed for image segmentation and lane detection tasks in this work. 
Our main contributions are as follows:
\begin{itemize}
    \item We introduce a novel DONN architecture for all-optical RGB images processing, demonstrating with image segmentation and lane detection tasks in autonomous driving. The architecture is designed with three separate channels, each dedicated to processing the red ("R"), green ("G"), and blue ("B") components of an RGB environment input. We incorporate optical skip connections between the early layers and the predication layers within each channel to address the vanishing gradient problem and achieve effective model training.

    \item We provide a comprehensive analysis of the proposed model for the image segmentation task using CityScapes dataset\cite{cordts2016cityscapes}. Our results show that our architecture delivers more detailed and accurate segmentation compared to the existing DONN system~\cite{lin2018all}. 

    \item We evaluate the model's performance in lane detection tasks through two case studies: track trace extraction for robotic cars in an indoor playground, and lane extraction for urban scenes simulated in CARLA\cite{dosovitskiy2017carla}. Additionally, we demonstrate the model's generalizability across different maps, weather conditions, and times of the day.  

\end{itemize}

%In this work, we propose a novel DONN architecture targeting image segmentation and lane detection tasks in autonomous driving. Specifically, the model is designed with three channels, each dedicated to processing the "R", "G", and "B" channels from camera inputs separately. For each channel, we incorporate an optical skip connection between early layers to the predication layer to solve the vanishing gradient problem for effective model training. To evaluate the model, we \textbf{first} present the comprehensive analyses of the proposed model on the image segmentation task using CityScapes dataset\cite{cordts2016cityscapes}, resulting in more detailed segmentation when compared with the single-channel DONN system. \textbf{Second}, we highlight the performance of the model in lane detection tasks through two case studies: (1) track trace extraction for robotic toy cars in an indoor playground, and (2) lane extraction for urban scenes simulated in CARLA, where we further demonstrate the model's generalizability across different maps, weather conditions, and times of the day.  

\section{Background and Related Works}
\label{sec:background}

\subsection{Deep Learning for Autonomous Driving}
\label{sec:auto_drive}

{Autonomous driving (AD) has been a prominent research topic in recent years, and it gets lots of attention and practical applications through the use of deep learning systems. There are mainly four steps involved in enabling autonomous driving: (1) Sensor data acquisition and fusion, where the vehicle's sensors continuously capture data from the surrounding environment. The vehicle is typically equipped with multiple kinds of sensors, such as cameras, LiDAR, radar, and GPS, to capture comprehensive environment information. The sensor data is then fused to generate an accurate model of the environment; (2) Perception, where deep learning models are employed to interpret the surrounding environment. Key tasks for the perception model include object detection, semantic segmentation, and lane detection; (3) Predictions and planning, where the vehicle predicts the future movements of other agents and plans its own movement to ensure safe driving; (4) Control generation, where the control signal is generated to physically move the vehicle along the planned path. 
The ultimate goal for autonomous driving is to realize safe, reliable, and energy-efficient driving that outperforms human drivers in all environments and conditions\cite{wu2017squeezedet}. 
 
Developing high-fidelity and generalizable deep learning models for perception tasks is critical in autonomous driving assistance systems \cite{rizzoli2022multimodal}. The first deep learning models used for semantic segmentation perform pixel-wise classification on the input image based on the feature extraction of the input image using convolutional models, including Fully Convolutional Network (FCN) \cite{long2015fully}, VGG \cite{simonyan2014very}, ResNet \cite{he2016deep}, and MobileNet \cite{howard2017mobilenets}. Novel network architectures designed specifically for semantic segmentation are then proposed. For example, ParseNet \cite{liu2015parsenet} adds global context to the deep learning network, and DeconvNet \cite{noh2015learning} incorporates deconvolution and unpooling layers to improve the performance. Recently, attention-based models have gained significant attention. Based on the transformer architecture \cite{vaswani2017attention}, the Vision Transformers (ViT) \cite{dosovitskiy2020image} is proposed for image understanding, which presents a convolution-free, transformer-based vision approach. 

%Besides RGB images, more and more sensor data is involved for more accurate and efficient segmentation tasks. For example, with LiDAR data, PointNet\cite{} is proposed for general-purpose 3D pointcloud segmentation. RandLANet\cite{} and Cylinder3D\cite{} are proposed specifically to handle the sparse nature of LiDAR samples better. With the growth in the availability of heterogeneous data, the AD system will acquire more detailed, accurate, and comprehensive raw data by sensing the surrounding environment, while posing more and more significance and challenges on the perception models and algorithms used in the AD system.  

In autonomous driving, perception models must not only be accurate but also maintain low inference times to ensure real-time responses. Additionally, as the deep learning model is deployed on edge computing platforms in autonomous vehicles that are resource-constrained, energy efficiency of the model is crucial for practical deployment.
However, existing digital deep learning models are complex and requires heavy computation involving millions of neurons. This results in high power consumption and expensive memory requirements, posing significant challenges for efficient deployment.

}

\subsection{Optical Computing Systems}
\label{sec:onn}
In the past decade, there has been an increasing interest in optical computing platforms because of their potential to realize fast, massively parallel computation at low power consumption~\cite{solli2015analog}. There are two main directions to bring the advantages of optics into the computing paradigm: (1) integrated photonics-based optical systems, which integrate optical computing units with digital systems to accelerate the computing speed while reducing the energy cost. This approach requires massive conversions between digital and analog signal for information inputs, computing, and outputs; (2) free-space diffractive optical systems, which have direct access to all optical degrees of freedom that carry information of the analog input wave without the requirement of analog-to-digital conversions (ADCs), further saving the energy cost while maintaining more accurate input information. 

\begin{figure}
    \centering  \includegraphics[width=1\linewidth]{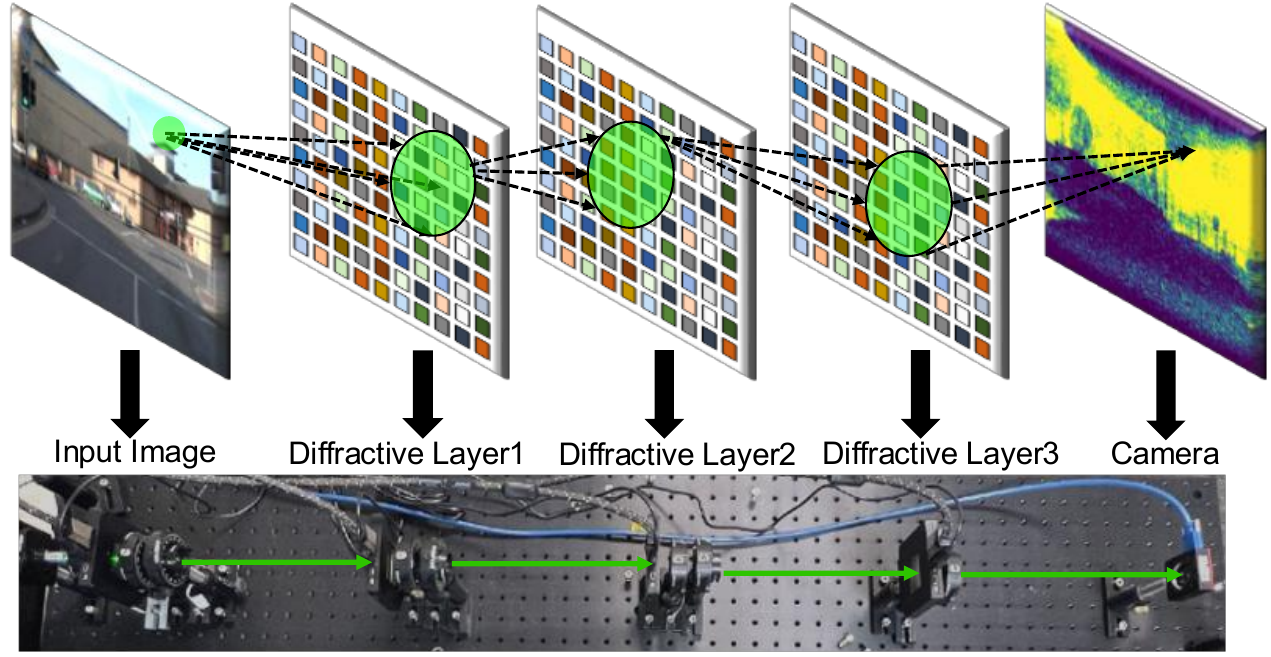}
    \caption{The basic DONN framework. The upper part illustrates the extracted model for a DONN system including the input image, three diffractive layers, and the system output for the image segmentation task. The lower part shows the hardware deployment for the all-optical inference with a trained DONN system. It includes a laser source for input image encoding and computation, three SLMs for diffractive layers, and a camera to capture the system output. }
    \label{fig:donn_system}
\vspace{-5mm}
\end{figure}

\subsubsection{Integrated Photonics-based Optical Processor }

Integrated photonics-based processors~\cite{shen2017deep} are designed and implemented with on-chip interferometers and waveguide-embedded light sources to replace or empower their electronic counterparts. They work as an efficient on-chip task-specific accelerator in a whole computing system with their reconfigurability and ease of integration with electronics, making them powerful and versatile for optoelectronic computing. 

However, the computing systems integrated with photonic devices often require pre-processing of information with ADCs/DACs, which can be energy-costly. In the optoelectronic computing system, nearly 50\% of power is from electrical memory, and ADCs/DACs take another 20-30\%, while the photonic circuit consumes less than 10\% total power~\cite{ramey2020silicon}. The speed and power consumption for memory and ADCs/DACs determine the overall benefits the computing system could gain from the integration with optical accelerators.

\subsubsection{Free-space Diffractive Optical Neural Networks (DONNs)}

Unlike digital NNs and integrated photonics-based optical computing systems, DONNs take the direct optical information as system inputs for computing without requiring input pre-processing, offering an efficient solution to further improve the perception performance for autonomous driving.
The DONN system performs all-optical inference for ML tasks based on physical phenomena by manipulating the light signal's features, such as phase and amplitude, with passive optical devices. Thus, the DONN system features high system throughput, high computation speed, while having low power consumption~\cite{lin2018all, shen2017deep, hughes2018training}. 
As shown in Figure \ref{fig:donn_system}, in the DONN system, the information is encoded onto the light signal emitted by a laser source, through modulation methods. Multiple diffractive layers are employed to manipulate features of the light signal, such as its phase and amplitude. These layers are implemented with spatial light modulators (SLMs). The layers form the neural network and each pixel in the layer acts as a neuron, which mimic the morphology of artificial neural network with passive optical devices, while requiring no additional energy for computation \cite{mengu2019analysis, mengu2020scale}. At the end of the system, a detector is used to capture the results. Compared to digital NNs, the DONN system encodes weights as complex-valued transmission coefficients (phase modulation to the light signal) within the diffractive layers. The free-space light diffraction provides neuron connectivity across neighboring layers with forward propagation within the DONN system. 

%As a result, the DONN system can be an economic, eco-friendly and low-latency solution for perception tasks in autonomous driving, owing to the following reasons: (1) The DONN system saves the expensive and error-prone ADC for input sensing. The system interacts with the analog signal, i.e., the information-encoded light signal, directly for the computation. (2) The computation and data movement occur at the speed of light in the DONN system. Computation is achieved through phase modulation of the light signal, and the data movement is realized by light diffraction between layers. (3) No additional energy is required during all-optical inference beyond the laser source used for information encoding. In the DONN system, the computation (phase modulation) is performed by passive optical devices. Once passive optical devices are fabricated and assembled, they provide modulation to the light signal at no extra energy cost. The data movement (light diffraction) occurs naturally through light propagation process at no extra cost. 

Existing DONN systems are typically implemented with single channel, restricting their applications to grayscale input images only \cite{lin2018all,sun2023review}. Additionally, existing DONN systems are primarily used for classification tasks with one-hot represented labels \cite{lin2018all,li2020multi,zhou2021large,yan2022all}, limiting applications to more complex image processing tasks. In this work, we propose a novel DONN architecture designed for advanced RGB image processing tasks including image segmentation and lane detection. Our framework takes an image as input and produces a processed image as output, expanding the functionality of DONNs beyond simple classification.

\section{Approach}
\label{sec:approch}

\begin{figure*}
    \centering
    \begin{minipage}{0.68\textwidth}
        \centering
        \includegraphics[width=\textwidth]{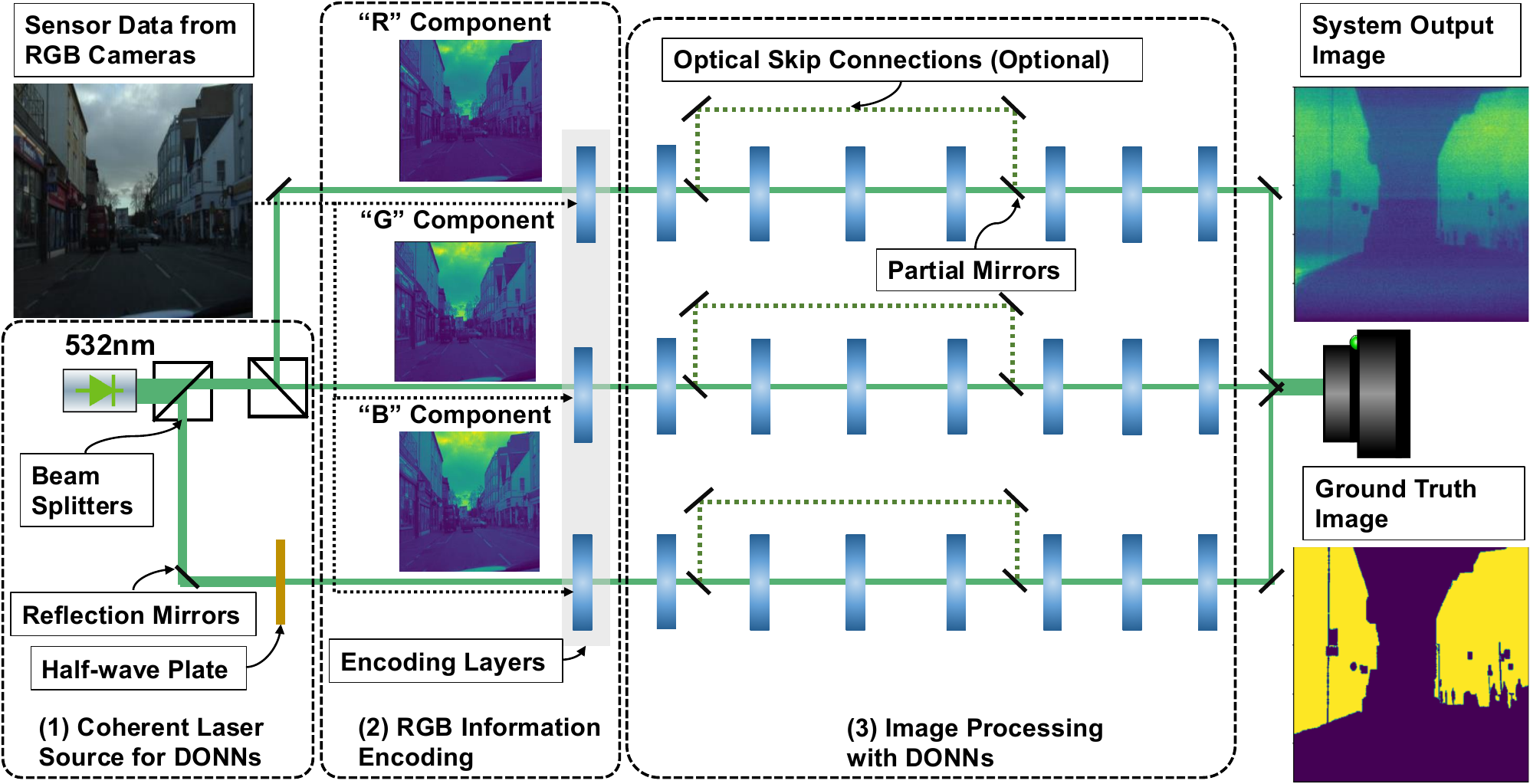}
        \subcaption{Illustration of the DONN system designed for RGB image processing. It mainly includes three parts: (1) Coherent laser source emitted the light signal. (2) Information coding, where the light information ('R', 'G', 'B' components) captured by the front passive optical system in the RGB camera is used to generate the encoding layers for three separate DONN channels. When the light signal propagates through the encoding layer, the corresponding information will be encoded on the light signal. (3) Image processing computation with trained diffractive layers. We can implement optical skip connections for deep DONN systems.  }
        \label{fig:model_structure}
    \end{minipage}%
    \hfill
    \begin{minipage}{0.31\linewidth}
        \centering
        \includegraphics[width=0.83\linewidth]{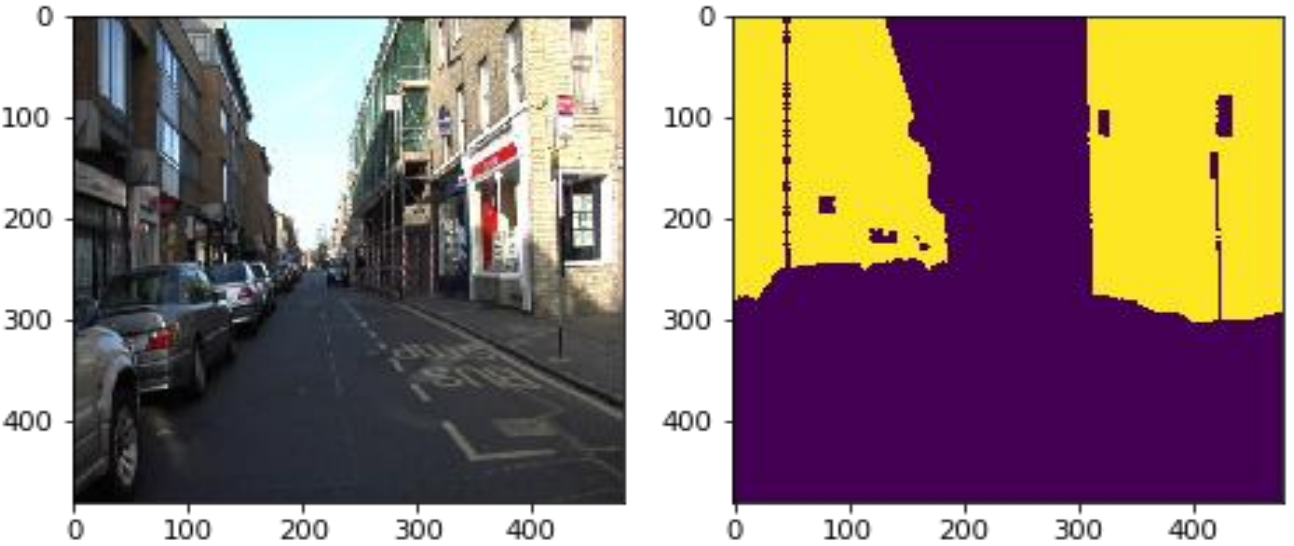}
        \subcaption{The input image and the ground truth image from CityScapes.}
        %\vspace{-5pt}  % Add space between (b) and (c)
        \label{fig:data_city}
        
        \includegraphics[width=0.83\textwidth]{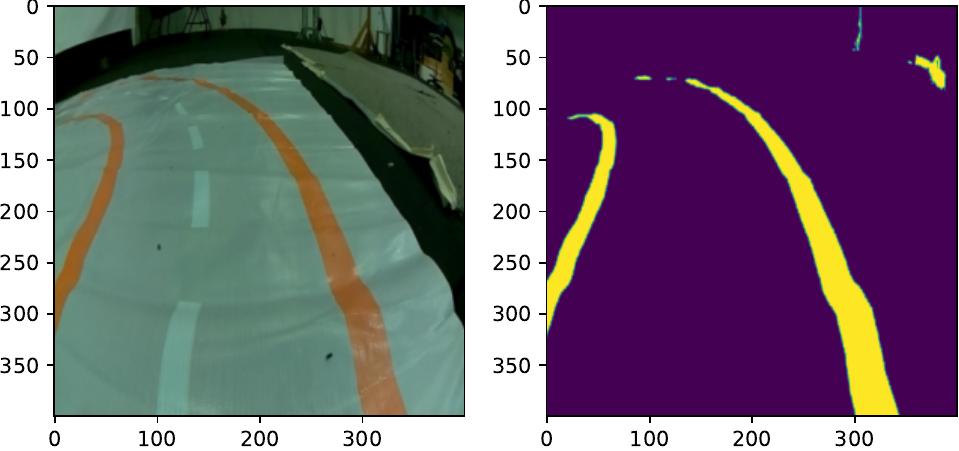}
        \subcaption{The input image and the ground truth image from customized indoor track.}
        \label{fig:data_toy}
        
        \includegraphics[width=0.83\textwidth]{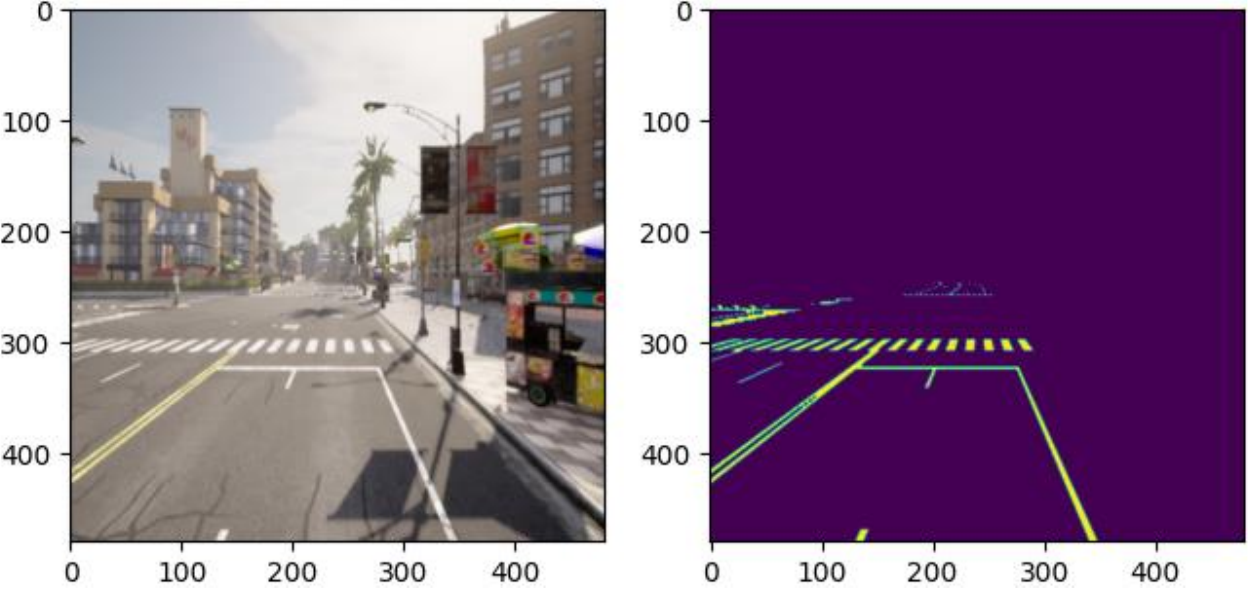}
        \subcaption{The input image and the ground truth image from simulations in CARLA.}
        \label{fig:data_carla}
    \end{minipage}
    \caption{Illustrations of the DONN system for image segmentation and lane detection with RGB images.}
    \vspace{-3mm}
\end{figure*}

In this section, we introduce a novel DONN architecture to process RGB images for semantic image segmentation.  
To train a DONN model for machine learning tasks, the optical responses during the all-optical inference by DONN systems is first emulated on digital platforms with numerical modeling. The training process leading the design of a functioning DONN system is performed on digital platforms with the mathematical approximation of the system.
%The trainable parameters in the system is the phase modulation provided at each diffractive layer, where by manipulating the phase modulation, the final diffraction pattern is changed accordingly, and the final prediction is produced by reading the final diffraction pattern from the camera. Hence, similar to the training process of conventional DNNs, optimal weights for phase modulation in diffractive layers in DONN can be obtained by minimizing the commonly used machine learning loss function~\cite{lin2018all}.

\subsection{Numerical Modeling of DONN Systems for RGB Images}
\label{sec:donn}

The model structure is illustrated in Figure \ref{fig:model_structure}. Targeting RGB images, we build three channels in the DONN system for "R", "G", and "B" image components respectively. 
\textbf{(1)} The "R", "G", and "B" image components of the RGB input are obtained after the optical sensor captures incoming light and the optical passive filter spatially separates it into red ("R"), green ("G"), and blue ("B") components. The sensed data is used to configure/fabricate the "Encode Layers" for each component in each channel. 
\textbf{(2)} %For each channel in our DONN system, the input is a grayscale image corresponding to "R" component image, "G" component image, and "B" component image, respectively. 
The three grayscale input image information for each component is then encoded on the coherent laser light signal. The laser beam is splitted into three sub-beams with same intensity by beam splitters, where the half-wave plate is used to decay the intensity of one of the sub-beams by half for even light intensity distribution over all channels. The image information for each component is encoded on the laser beam for each channel with "Encoding Layers". 
{\textbf{(3)} For each channel, there are separate diffractive layers to manipulate the information-encoded light signal within the channel. The light signal is diffracted in the free space between diffractive layers, and modulated via phase modulation at each layer. 
\textbf{(4)} Finally, the diffraction patterns after light propagation regarding light intensity distribution from all three channels are mixed and captured at the detector plane as the processed image output by the DONN system. %The modeling for each channel is following the same emulation methods while the phase modulations in channels are training separately. 
\subsubsection{\textbf{Forward Function for DONN Systems}}

For each channel, the input information (e.g., "R" component image) is encoded on the coherent light signal, where the light signal is expressed with complex-valued numbers in the DONN emulation. Its wavefunction after the encoding layer is expressed as $f^{0}(x_{0}, y_{0})$. The wavefunction after light diffraction over diffraction distance $z$ to the first diffractive layer can be seen as the summation of the outputs at the input plane, i.e., 
\begin{equation}
\label{eq:diffraction_time}
    f^{1}(x, y) = \iint f^{0}(x_{0}, y_{0})h(x-x_{0}, y-y_{0}, z)dx_{0}dy_{0}
\end{equation}
where $(x, y)$ is the coordinate on the receive plane, and $h$ is the impulse optical response function of free space, which is the mathematical approximation for light diffraction, e.g., Rayleigh-Sommerfeld approximation, Fresnel approximation, Frauhofer approximation~\cite{tobin1997introduction}, resulting in the non-trainable parameters in DONN systems. 
For example, in this work, we use Fresnel approximation for DONN emulation, where 
\begin{equation}
\label{eq:diffraction_time_end}
    h(x, y, z) = \frac{\exp(ikz)}{i\lambda z}\exp\{\frac{ik}{2z}(x^{2} + y^{2})\}
\end{equation}
where $i=\sqrt{-1}$, $\lambda$ is the wavelength of the laser source, $k=2\pi/\lambda$ is free-space wavenumber.

Equation \ref{eq:diffraction_time} can be calculated with spectral algorithm, where we employ Fast Fourier Transform (FFT) for fast and differentiable computation, i.e., 
\begin{equation}
\label{equ:diffraction_freq}
    U^{1}(\alpha, \beta) = U^{0}(\alpha, \beta)H(\alpha, \beta, z)
\end{equation}
where $U$ and $H$ are Fourier transformations of $f$ and $h$, respectively. 

After light diffraction, the resulting wavefunction $U^{1}(\alpha, \beta)$ in Equation \ref{equ:diffraction_freq} is transformed to time domain with inverse FFT (iFFT) for phase modulation, and the phase modulation $W(x, y)$ provided by the diffractive layer is applied to the light wavefunction by matrix multiplication, i.e.,
\begin{equation}
\label{eq:phase_mod}
    f^{2}(x, y) = \text{iFFT}(U^{1}(\alpha, \beta)) \times W_{1}(x, y)
\end{equation}
where $W_{1}(x, y)$ is the phase modulation in the first diffractive layer, $f^{2}(x, y)$ is then the input light wavefunction for the light diffraction for the second diffractive layer. 

We wrap one computation round of light diffraction and phase modulation at one diffractive layer as a computation module named \textbf{DiffMod}, i.e.,
\begin{equation}
\label{eq:diffmod}
    \text{DiffMod}(f(x, y), W) = L(f(x, y), z) \times W(x, y)
\end{equation}
where $f(x, y)$ is the input wavefunction, $W(x, y)$ is the phase modulation, $L(f(x, y), z)$ is the wavefunction after light diffraction over a constant distance $z$ in time domain.  

Thus, in a multiple diffractive layer constructed DONN system, the forward function $f^{*}$ can be computed iteratively for the stacked diffractive layers. For example, for a 3-layer constructed channel in the system, the forward function is
\begin{equation}\label{eq:forward}
    \begin{split}
       f^{*}(f^{0}(x, y), W) = \text{DiffMod}(\text{DiffMod}(\text{DiffMod}(f^{0}(x, y), \\
       W_{1}(x, y)),W_{2}(x, y)), W_{3}(x, y)) \\
    \end{split}
\end{equation}

\subsubsection{\textbf{Optical Skip Connections Between Diffractive Layers}}

Adopted from residual neural networks~\cite{he2016deep}, we implement the optical skip connection into the DONN system to recover the vanishing gradients for effective model training in deep systems. \textbf{However, the optical skip connection involves free-space light diffraction during the propagation through the skip distance between layers.}

As shown in Figure~\ref{fig:model_structure}, an optical skip connection is implemented after the first computing layer and before the fifth computing layer in (3). The distance between each layer is $z$, the propagation distance for the optical skip connection is thus $4\times{z}$. When the wavefunction of the light signal after propagating through the first layers is $f^{*}_{1}$ and the wavefunction of the light signal after the fourth layers is $f^{*}_{4}$, the wavefunction of the input light signal to the fifth layer $f^{0}_{5}$ is 
\begin{equation} \label{equ:optical_skip}
    f^{0}_{5} = f^{*}_{4} + L(f^{*}_1, 4 \times z)
\end{equation}
Thus, the forward function with the optical skip connection in Figure \ref{fig:model_structure} is 
\begin{equation}
\label{eq:forward_skip}
\begin{split}
    f^{*}(f_{0}(x, y), \mathbf{W}) = \text{DiffMod}(\text{DiffMod}(\text{DiffMod}(\text{DiffMod} \\ ((\text{DiffMod}  (\text{DiffMod}(\text{DiffMod}(f_{0}(x, y), \mathbf{W_{1}}), \mathbf{W_{2}}), \mathbf{W_{3}}) \\  \mathbf{W_{4}}) {\color{blue}{+ L(f^{*}_{1}, 4\times z)}}, \mathbf{W_{5}}), \mathbf{W_{6}}), \mathbf{W_{7}}) 
\end{split}
\end{equation}
Note that we can insert optical skip connections between arbitrary layers while setting the corresponding skip distance for light diffraction. The optical skip connections can be implemented with passive optical devices including partial mirrors and reflection mirrors as shown in Figure~\ref{fig:model_structure}, which requires no extra energy cost.

\subsubsection{\textbf{Light Intensity at the Detector}}

The wavefuction $f^{*}$ is a complex-valued number and it can be represented by $f^{*} = A + iB$, where $A$ and $B$ are real numbers. Thus, the output intensity pattern is expressed as 
\begin{equation}\label{eq:intensity}
    I = A^{2} + B ^{2}
\end{equation}

The modeling for each channel is following the same numerical emulation as shown in Equation \ref{eq:forward_skip} and \ref{eq:intensity}. The forward function $I$ for the "R", "G", "B" channels can be expressed with $I_{R}$, $I_{G}$, $I_{B}$, where the inputs $f^{0}(x, y)$ are the corresponding input $f^{0}_{R/G/B}(x, y)$ from the 'R/G/B' component, and $W$ are the corresponding phase modulations $W_{R/G/B}$ trained separately in each channel. The final diffraction pattern at the detector of this three-channel RGB-image segmentation DONN framework is expressed as 
\begin{equation}\label{eq:forward_all}
    I_{det} = I_{R} + I_{G} + I_{B}
\end{equation}
where $I_{det}$ is the intensity pattern captured at the detector. 

\subsection{Dataset Processing}\label{sec:dataset}
In this work, the ground truth images in datasets are preprocessed as binary images for image segmentation task and lane detection task.
%We target (1) the binary segmentation task on RGB images, where each pixel is labeled as either ”object” ("1"s) or ”non-object” ("0"s); (2) the binary lane detection task on RGB images, where the lane markers and road tracks are labeled as "1"s while others are labeled as "0"s. 
Specifically, we utilize three datasets in this work: (1) CityScapes dataset~\cite{cordts2016cityscapes}, which captures complex urban scenes and is used for segmentation tasks. In this work, we modify its labels by making it binary such that building pixels are marked as ”1”s and non-building pixels are marked as ”0”s, as shown in Figure \ref{fig:data_city}. (2) Indoor track detection dataset. We record the video with a robotic car driving along a real-world indoor track. Each frame is extracted from the video and manually annotated to build the dataset. As shown in Figure \ref{fig:data_toy}, we annotate the tracks as "1"s, while others are marked as "0"s. (3) Simulated driving dataset. This dataset includes simulated urban driving scenes generated in CARLA~\cite{dosovitskiy2017carla}. As shown in Figure \ref{fig:data_carla}, the pixels for lane markers on the road are set as "1"s while others are set as "0"s.

\subsection{Model Training}
\label{sec:train}

The all-optical DONN system is trained with the numerical modeling as shown in Equation \ref{eq:forward_skip}, \ref{eq:intensity}, and \ref{eq:forward_all} on digital platforms. The trainable parameters in the system are the phase modulation $W$ provided at each diffractive layer. By manipulating the phase of the information-encoded light signal at each diffractive layer during light propagation, the final diffraction pattern changes accordingly, and the system output is produced by reading the final intensity pattern on the camera. Hence, similar to the training process of conventional DNNs, optimal phase modulation in diffractive layers in DONN can be obtained by minimizing the commonly used machine learning loss functions~\cite{lin2018all}.

The ground truth of the segmented image is first processed as binarized images as described in Section \ref{sec:dataset}. The intensity distribution for the ground truth image is $I_{GT}$ as shown in Figure \ref{fig:data_city} to \ref{fig:data_carla}. %For example, in Figure \ref{fig:data_city}, the image segmentation task with CityScapes dataset, the object is 'buildings' and the pixels for building in the image are set as '1's while other pixels are set as '0's. For the lane detection task with the simulation sample from CARLA in Figure \ref{fig:img_proc_ours}, the pixels for lanes shown on the road are set as '1's while others are set as '0's.  
With the output from the DONN system expressed with Equation \ref{eq:forward_all}, we implement the loss function between the ground truth image ($I_{GT}$) and the system output $I_{det}$. Then, we apply Mean Square Error (MSE) loss between $I_{GT}$ and $I_{det}$, the loss function $L$ is calculated as
\begin{equation}\label{equ:mse_loss}
    \mathcal{L}_{\text{MSE}} = \frac{1}{N^{2}} \sum_{i=1}^{N} \sum_{j=1}^{N} (I_{GT}^{ij} - I_{det}^{ij})^2
\end{equation}

Moreover, we can implement different loss functions for the training such as Binary Cross-Entropy (BCE) loss and Dice loss~\cite{milletari2016v}. The explorations regarding different training loss functions are shown in Section \ref{sec:seg_city_loss}. Thus, the whole system is designed to be differentiable and compatible with conventional automatic differential engines. %The numerical modeling of the DONN system described above has been physically verified by \cite{zhou2021large, chen2022physics}.

\section{Experiments}
\label{sec:experiments}

This section presents the evaluations of the proposed DONN system for image segmentation task and lane detection task in autonomous driving. First, Section \ref{sec:seg_city} provides the evaluations on the image segmentation task with the CityScapes dataset, including model explorations and the performance comparisons with existing image segmentation systems~\cite{lin2018all, ronneberger2015u}. Then, the model is evaluated with lane detection task in Section \ref{sec:lane} with the customized indoor track dataset and the simulated driving dataset with CARLA. The generalizability of the model is also explored with different environment conditions with the dataset from CARLA.

\subsection{Experiments Setups}

Our DONN systems are implemented with three channels for RGB input images. For different tasks with different datasets, the DONN system is prepared with different setups. For the image segmentation task with CityScapes dataset in Section \ref{sec:seg_city}, where the input samples are more complex, we set system size as $480 \times 480$. The comparison and explorations are conducted with the system implemented with fifteen diffraction layers and three optical skip connections between the first layer and the fifteenth layer, the second layer and the fourteenth layer, and the third layer and the thirteenth layer for each separate channel. 
For the lane detection task with the indoor customized track dataset in Section \ref{sec:lane}, where the input samples are clean and minimalistic, featuring a well-defined object against a plain background, the input image and the system size is set as $400 \times 400 $. The DONN system is implemented with eight diffraction layers and three optical skip connections between the first layer and the sixth layer, the second layer and the seventh layer, and the third layer and the eighth layer, for each each channel. 
The same system architecture is used for lane detection with the simulated driving dataset with CARLA in Section \ref{sec:lane}, while the system size is set to $480 \times 480$.
The Intersection over Union (IoU) is used as the metric to evaluate the segmentation performance. The IoU is calculated between the binary ground truth image and the binarized output image from the DONN system. 

The input information is encoded on the coherent light signal with the wavelength of $532~nm$. The pixel size for each diffractive layer is $36~um$. The physical distances between layers, first layer to source, and final layer to detector, are set as $27.94~cm$, i.e., $z = 27.94~cm$. A CMOS camera is placed at the end of the system to capture the output image. In this work, the loss function is formulated with MSE loss shown in Equation \ref{equ:mse_loss}. For each task, the model is trained with $500$ epochs with batch size of $64$. All implementations are constructed using PyTorch v1.8.1. The experiments are conducted with Nvidia RTX A6000, while the simulated driving dataset with CARLA is collected with Nvidia 4090 Ti GPU.

\subsection{Image Segmentation with CityScapes}
\label{sec:seg_city}

%2975 training, 500 val
% the training model for this: python3 auto_inf.py --vis True --whether-load-model True --depth 15 --sys-size 480 --model-save-path ./model_city_build_road_rgb_15L/ --start-epoch 791 --image-save-path ./vis_city_build_road_rgb_15L/ --datapath ./cityscapes_process_road/ --batch-size 50
% binarized with more delicated denoising
% the inference IoU is 0.70

% for picked dataset1: python3 auto_inf_dataset1.py --vis True --whether-load-model True --depth 8 --sys-size 480 --model-save-path ./model_city_dataset1/ --start-epoch 801 --image-save-path ./vis_city_dataset1_0318/ --datapath ./dataset1/ --batch-size 1
%binarized with mean() * 0.9
% the iou for train is 0.45 (367 figures), but it has very nice figures
% the iou for inference is 0.42 (101 figures)

% for gray-scale images: python3 auto_gray.py --depth 15 --sys-size 480 --datapath ./cityscapes_process_road/ --model-save-path ./model_city_build_road_gray_15L/ --image-save-path ./vis_city_build_road_gray_15L/ --device cuda:1 --epochs 802 --batch-size 64 --result-record-path ./log_city_road_gray_15L.csv --lr 0.15 
% same binarized method, the inference iou is 0.36

% for bce training: 

% for carla sim: python3 auto_inf_carla.py --whether-load-model True --vis True --depth 8 --sys-size 480 --model-save-path ./model_sim_0317/ --image-save-path ./vis_sim_0320/ --start-epoch 291 --batch-size 1 --datapath ./carla_sim_0317

This section provides the evaluation for the DONN system using CityScapes dataset, which contains $2975$ training samples and $500$ evaluation samples. As shown in Figure \ref{fig:data_city}, the input images in CityScapes are rich in environmental information including pedestrians, buildings, vehicles, and other urban elements. The DONN system is implemented with the system size of $480 \times 480$. The impact of the number of layers on segmentation performance is first explored in Section \ref{sec:seg_city_perf}. 
Further, the effects of different loss functions are explored in Section~\ref{sec:seg_city_loss}, and the performance comparison with existing segmentation systems are provided in Section~\ref{sec:seg_city_gray} and Section~\ref{sec:seg_city_unet}. These evaluations are conducted using a 12-layer DONN system with three optical skip connections between the first three and last three layers.

\subsubsection{\textbf{Image Segmentation Performance}}
\label{sec:seg_city_perf}

%Show the comparison results with single channel gray-scale system, both with optical skip connections

%add the IoU as metrics? 

%show the loss curve and inference curve here
\begin{figure}
    \centering
    \includegraphics[width=0.9\linewidth]{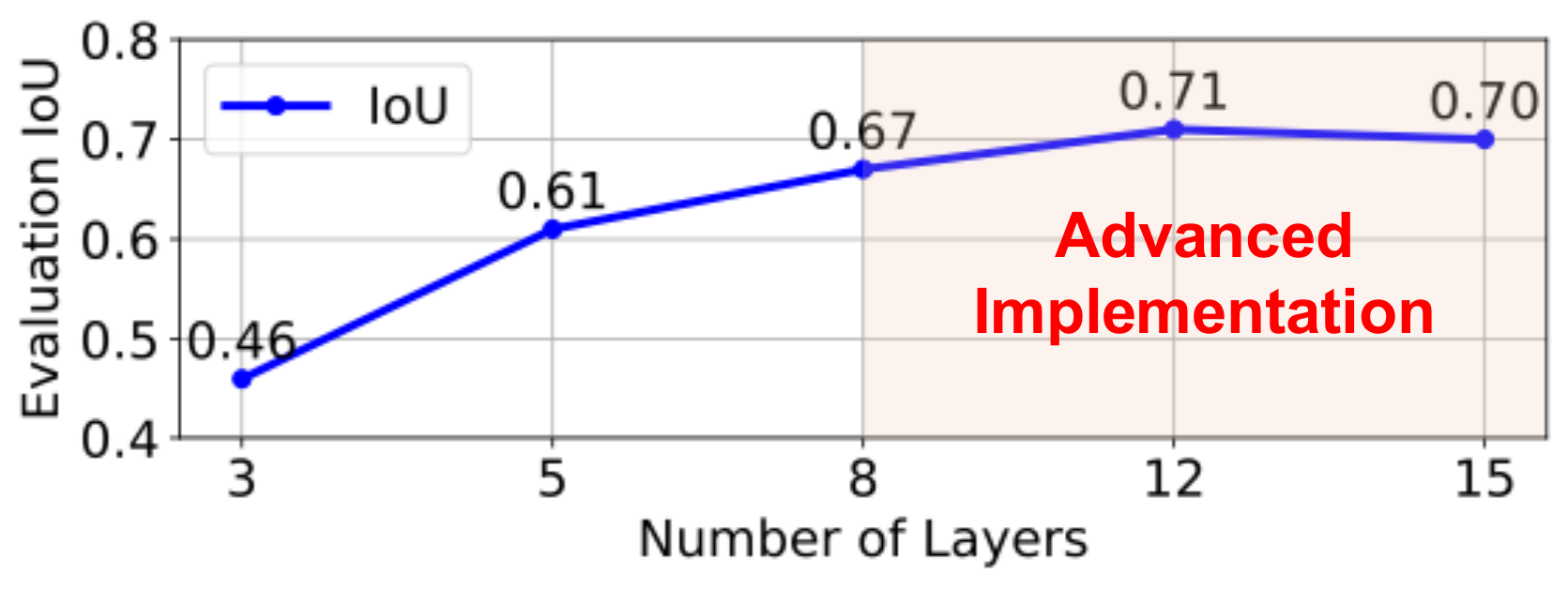}
    \caption{The segmentation performance (IoU) of DONN models with different numbers of layers.}
    \label{fig:acc_layer}
    \vspace{-2mm}
\end{figure}

This section presents the performance evaluation of the DONN system using RGB images from the CityScapes dataset. 
{Figure~\ref{fig:acc_layer} shows the segmentation performance for DONN systems with depths ranging from 3 to 15 layers. With a 12-layer DONN system, the evaluation IoU is $0.71$. While deeper architectures generally lead to improved performance, implementing DONN systems with more layers is challenging due to the exponential decay of light intensity for free-space propagation and the increased difficulty of optical alignment~\cite{lin2018all,fu2024optical}. 
Thus, more advanced optical devices and fabrication technology are required for stable and scalable hardware implementations of DONN systems, such as optoelectronic computing~\cite{zhou2021large} and metasurface-enabled on-chip integration~\cite{luo2022metasurface}.}

The normalized training loss curve and the evaluation IoU for a 12-layer DONN systems during the training process in shown in Figure \ref{fig:seg_city_curve}. The visualizations of inference samples are shown in Figure \ref{fig:city_seg_rgb} and Figure \ref{fig:discussion_samples}. Our model demonstrates strong confidence in segmenting prominent objects, such as buildings, from the sky and ground, when there is notable light contrast between the object and background. However, the model exhibits lower confidence in distinguishing fine details, such as segmenting vehicles from buildings. Additionally, some intricate features may be lost during the binarization process of the system outputs, potentially reducing segmentation precision.

%our DONN system implemented with 'R', 'G', and 'B' channels and trained with the MSE loss reports the evaluation IoU of $0.7$. %The normalized training loss curve and the evaluation IoU during the training process in shown in Figure \ref{fig:seg_city_curve}. 
%The visualizations of inference samples are shown in Figure \ref{fig:city_seg_rgb} and Figure \ref{fig:discussion_samples}. Our model demonstrates strong confidence in segmenting prominent objects, such as buildings, from the sky and ground, when there is notable light contrast between the object and background. However, the model exhibits lower confidence in distinguishing fine details, such as segmenting vehicles from buildings. Additionally, some intricate features may be lost during the binarization process of the system outputs, potentially reducing segmentation precision.

\begin{table}
\centering
\begin{adjustbox}{width=0.8\columnwidth, center}
\begin{tabular}{c|c|c|c|c}
\toprule
\bf{Input Image}     & RGB & RGB & RGB & Gray \\
\hline
\bf{Loss Function}   & MSE   & BCE & Dice & MSE\\
\hline
\bf{IoU}            & 0.70  & 0.66 & 0.66 & 0.36\\
\toprule
\end{tabular}
\end{adjustbox}
\caption{{Segmentation performance regarding IoU results on Cityscapes. We evaluate the model with different input image types and loss functions for comparison.}}
\label{tbl:city_iou}
\vspace{-3mm}
\end{table}

\begin{figure}
    \centering
    \includegraphics[width=0.9\linewidth]{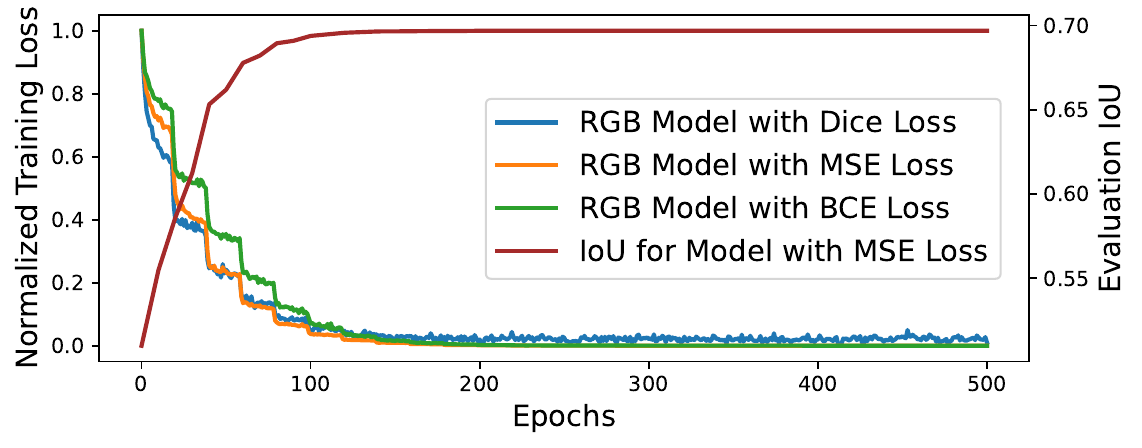}
    \caption{Normalized training loss and evaluation IoU curve during training process. }
    \label{fig:seg_city_curve}
\end{figure}

\begin{figure}[t]
     \centering
     \begin{subfigure}[b]{0.9\linewidth}
         \centering
         \includegraphics[width=1\linewidth]{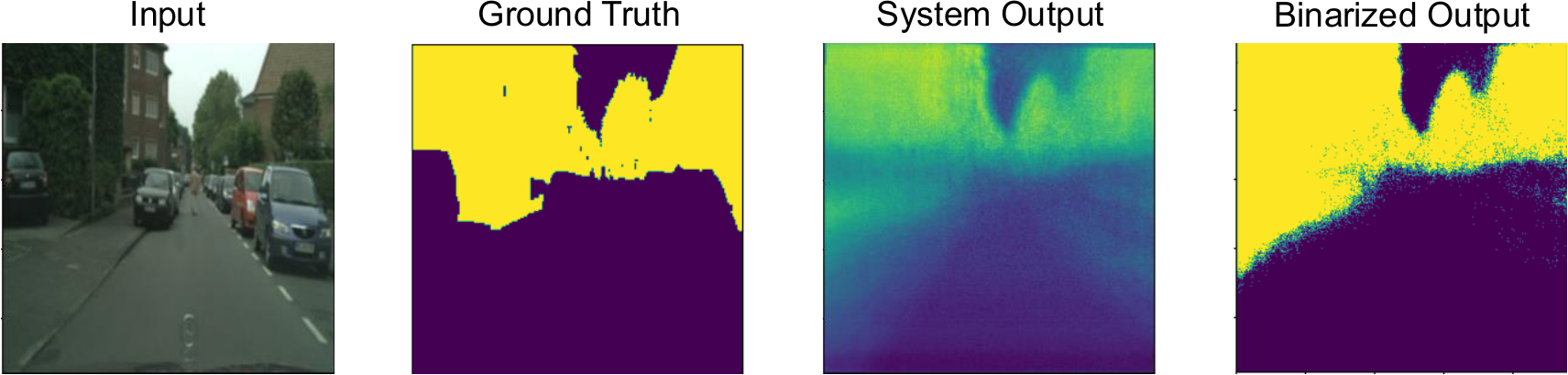}
         \caption{A visualization sample from the model trained with MSE loss.}
         \label{fig:city_seg_rgb}
     \end{subfigure}
     \hfill
    \begin{subfigure}[b]{0.45\linewidth}
         \centering
        \includegraphics[width=1\linewidth]{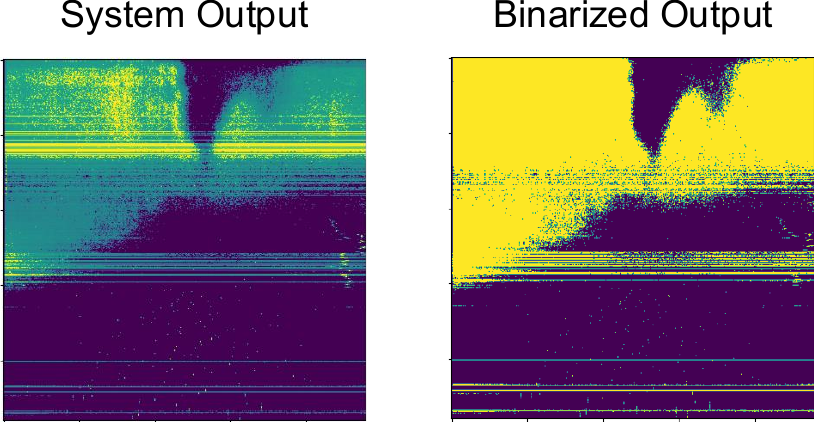}
         \caption{A visualization sample from the model trained with BCE loss.}
         \label{fig:city_seg_bce}
     \end{subfigure}
          \hfill
    \begin{subfigure}[b]{0.45\linewidth}
         \centering
        \includegraphics[width=1\linewidth]{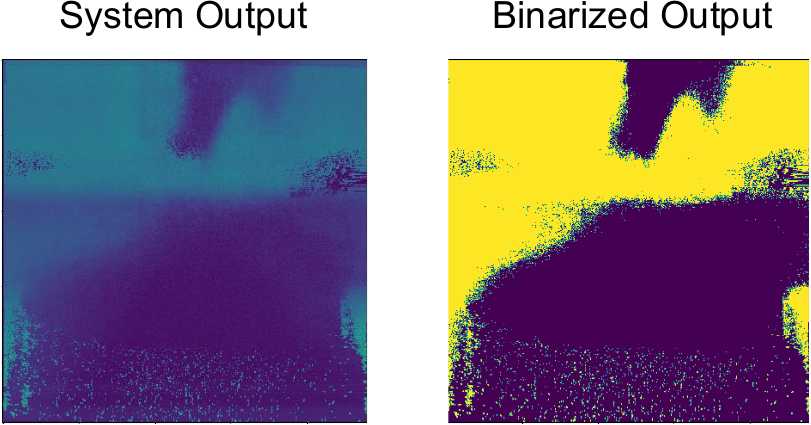}
         \caption{A visualization sample from the model trained with Dice loss.}
         \label{fig:city_seg_dice}
     \end{subfigure}

        \caption[Visualization samples for RGB-channel DONN system trained with different loss functions. ]{{Visualization samples for RGB-channel DONN system trained with different loss functions. Both direct outputs from the DONN system and the binarized outputs are shown for comparisons.}}
        \label{fig:seg_city_comp_loss}
\end{figure}

\subsubsection{\textbf{Loss Function Comparison}}
\label{sec:seg_city_loss}

Moreover, the model are trained with different loss functions including (1)~BCE loss, and (2)~Dice loss. The normalized training loss curves for models with both loss functions are shown in Figure \ref{fig:seg_city_curve}. The evaluation IoU with the model trained with BCE loss is $0.66$ while the evaluation IoU with the model trained with Dice loss is $0.66$, $4\%$ lower IoU than the model trained with MSE loss, as shown in Table~\ref{tbl:city_iou}. Additionally, the visualization sample for both models are provided in Figure \ref{fig:city_seg_bce} and Figure \ref{fig:city_seg_dice}. Compared to the system outputs from the model trained with MSE loss in Figure \ref{fig:city_seg_rgb}, the other two models shows model-specific noise such as  the horizontal noise line in Figure \ref{fig:city_seg_bce} with BCE loss, and the scatter noise cluster in Figure \ref{fig:city_seg_dice} with Dice loss.

%We train the 15-layer model with different loss functions including BCE loss and Dice loss. 
%in Equation \ref{equ:bce_loss}]
% and Dice loss. %in Equation \ref{equ:dice_loss}. %The normalized training loss curves for models with both loss functions are shown in Figure \ref{fig:seg_city_curve}. 
%As shown in Table \ref{tbl:city_iou}, the evaluation IoU for the model trained with BCE loss is $0.66$ while that for the model trained with Dice loss is $0.66$, $4\%$ lower IoU than the model trained with MSE loss. 
%Additionally, we provide the visualization sample for both models in Figure \ref{fig:city_seg_bce} and Figure \ref{fig:city_seg_dice}. Compared to the system outputs from the model trained with MSE loss in Figure \ref{fig:city_seg_rgb}, the other two models shows model-specific noise such as  the horizontal noise line in Figure \ref{fig:city_seg_bce} with BCE loss, and the scatter noise cluster in Figure \ref{fig:city_seg_dice} with Dice loss. However, the models trained with BCE loss and Dice loss demonstrates more confidence in classifying background pixels, i.e., pixels with '0', resulting in easier binarization. 

\subsubsection{\textbf{Single-Channel Model Comparison}}
\label{sec:seg_city_gray}

The RGB-channel system is further compared with the existing single-channel DONN system \cite{lin2018all}. The single-channel DONN system is implemented with fifteen diffractive layers and three optical skip connections between corresponding layers as our RGB-channel system, while the input image to the single-channel DONN system is a gray-scale image.   
With the same training setups, the IoU of the single-channel DONN system is $0.36$. A visualization sample is shown in Figure \ref{fig:city_seg_gray}. Compared to our RGB-channel DONN system, the single-channel system exhibits $35\%$ lower segmentation performance and shows less confidence during segmentation, as indicated by weaker light intensity contrast between the object and the background.

%We further compare the 15-layer RGB-channel system with the existing single-channel DONN \cite{lin2018all}. The single-channel DONN system is implemented with fifteen diffractive layers and three optical skip connections between corresponding layers as our RGB-channel system, while the input image to the single-channel DONN system is a gray-scale image.   
%With the same training setups, the IoU of the single-channel DONN system is $0.36$, as shown in Table \ref{tbl:city_iou}. A visualization sample is shown in Figure \ref{fig:city_seg_gray}. Compared to our RGB-channel DONN system, the single-channel system exhibits $34\%$ lower segmentation performance and shows less confidence during segmentation, i.e., weaker light intensity contrast between the object and the background.   
\begin{figure}[!t]
    \centering
    \includegraphics[width=1\linewidth]{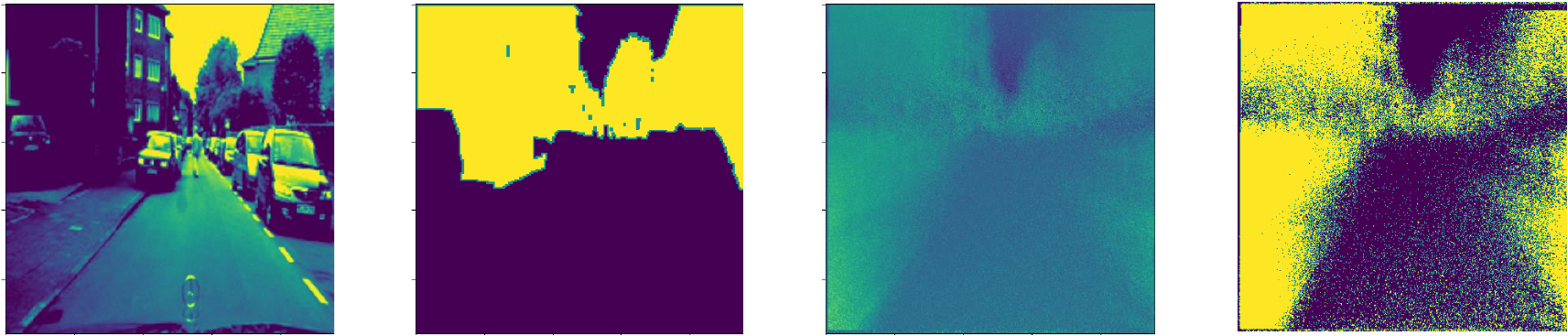}
    \caption{A visualization sample with gray-scale input image for a single-channel DONN system.}
    \label{fig:city_seg_gray}
\end{figure}

\paragraph{\textbf{Comparison with U-Net}}
\label{sec:seg_city_unet}

The RGB-channel DONN system is further compared with the digital U-Net~\cite{ronneberger2015u}. The CityScapes dataset is pre-processed with binary classes and the input size is set as $480 \times 480$ following the same dataset pre-processing as DONN systems.  
In the symmetric encoder–decoder structure of the U-Net architecture\footnote{https://github.com/deepmancer/unet-semantic-segmentation}, the contracting path (encoder) captures context through repeated blocks of 3×3 convolutions with ReLU activations and 2×2 max pooling with stride 2 for downsampling. The initial number of filters for the convolution layers is 32. At each downsampling step, the number of feature channels is doubled. The expansive path (decoder) restores spatial detail via transposed convolutions and integrates fine-grained features from the encoder through skip connections, which directly link corresponding resolution levels between the two paths. 
A final 1×1 convolutional layer maps the decoded feature maps to 2 output channels, generating the one-hot represented binary classes. The U-Net model is trained with BCE loss. 
The performance result is shown in Table~\ref{tbl:city_iou_unet}. The evaluation IoU for digital U-Net is $0.87$, while our DONN system achieves the evaluation IoU performance of $0.71$. The performance gap between digital neural networks and current DONN systems suggests that existing DONNs trade off model performance for energy efficiency, highlighting the need for more advanced algorithms and optical implementations. The visualization of a evaluation sample from U-Net is shown in Figure~\ref{fig:city_seg_unet}.

\begin{table}[!t]
\centering
\caption{{Segmentation performance regarding F1 score, precision, recall results on CityScapes, compared with U-Net.}}
\begin{tabular}{c|cccc}
\hline
   & \multicolumn{1}{c|}{F1 Score}  & \multicolumn{1}{c|}{Precision}  & \multicolumn{1}{c|}{Recall}  & IoU            \\ \hline
RGB-channel DONN system & \multicolumn{1}{c|}{0.83}  & \multicolumn{1}{c|}{0.79}  & \multicolumn{1}{c|}{0.88} & 0.71              \\ \hline
U-Net           & \multicolumn{1}{c|}{0.93} & \multicolumn{1}{c|}{0.93} & \multicolumn{1}{c|}{0.94} & 0.87           \\ \hline
\end{tabular}
\label{tbl:city_iou_unet}
\end{table}

\begin{figure}[!htbp]
    \centering
    \includegraphics[width=0.8\linewidth]{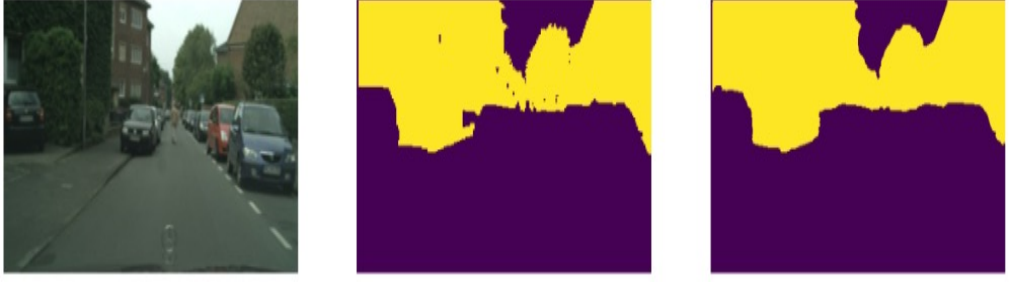}
    \caption[A visualization sample with U-Net model. ]{A visualization sample with U-Net model. The image from left to right corresponds to input image, groundtruth segmentation, and binarized system output.}
    \label{fig:city_seg_unet}
\end{figure}

\subsection{Lane Detection}
\label{sec:lane}
In this section, we evaluate the DONN model for the lane detection task. In Section \ref{sec:lane_toy}, we evaluate the model using a customized indoor dataset captured by a robotic vehicle (JETBOT\footnote{\url{https://jetbot.org/master/}}) navigating a real-world indoor track under uniform lighting conditions. A sample from the dataset is shown in Figure \ref{fig:data_toy}. Then, in Section~\ref{sec:lane_carla}, we evaluate the model with the simulated driving scenes using the CARLA simulator\footnote{\url{https://carla.org}}. We collect data across different simulated maps with different weather conditions and times of the day for comprehensive evaluations of the model's generalizability. A sample from the simulated driving dataset is shown in Figure \ref{fig:data_carla}. The RGB-channel DONN model used for lane detections with both datasets is implemented with eight diffraction layers and three optical skip connections between the first three layers and the last three layers.

\subsubsection{\textbf{Evaluations on Customized Indoor Track Dataset}}
\label{sec:lane_toy}

In this section, the model is evaluated with the customized indoor track dataset. The dataset is divided into training dataset with $4738$ images and evaluation dataset with $1000$ images. The RGB-channel DONN model is implemented with the system size of $400 \times 400$ and trained with MSE loss. 
The average IoU for the evaluation dataset is $0.80$. The visualized samples are shown in Figure \ref{fig:result_track_sample}, which shows that the proposed model can extract the track clearly in the indoor environment. 

\begin{figure}[!htb]
    \centering
\includegraphics[width=0.8\linewidth]{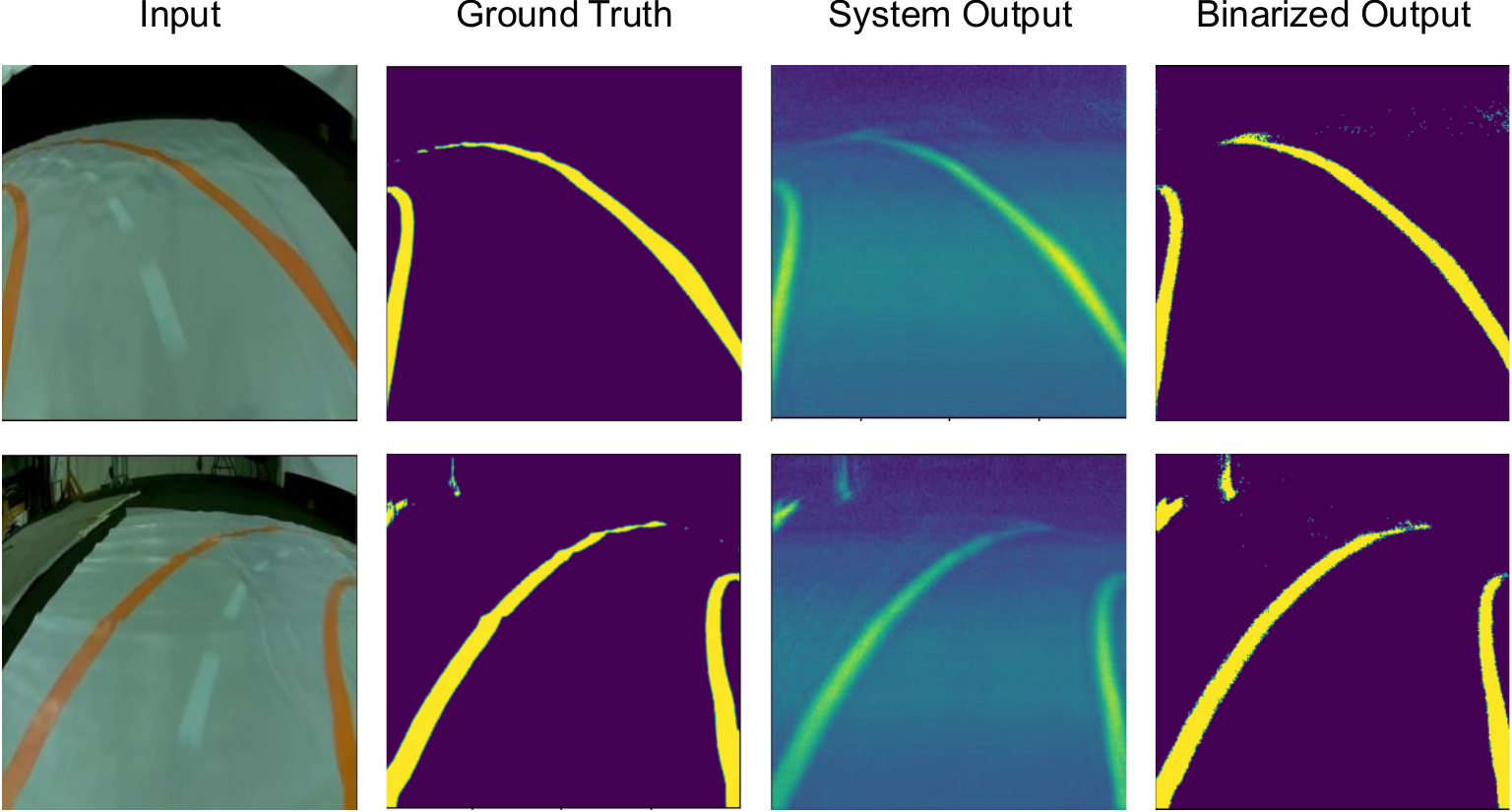}
    \caption{The inference samples for lane detection with the customized indoor track dataset.   }
    \label{fig:result_track_sample}
    \vspace{-3mm}
\end{figure}

\begin{figure*}
    \centering
    \begin{subfigure}[b]{0.865\textwidth}
         \centering
        \includegraphics[width=1\linewidth]{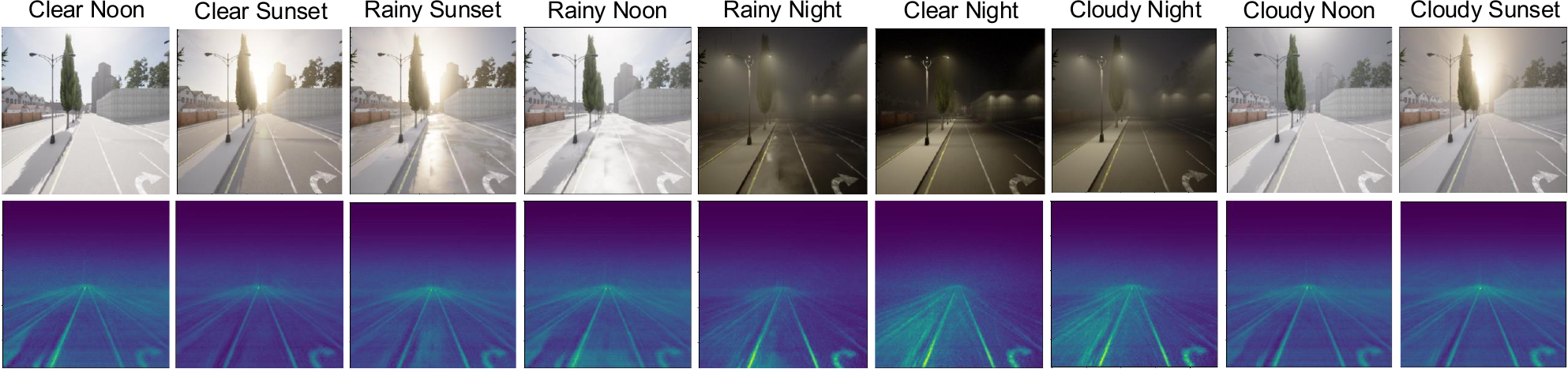}
         \caption{The system outputs for lane detection in 'Map01' under different environmental conditions.}
         \label{fig:gen_map01}
     \end{subfigure}
     \hfill
    \begin{subfigure}[b]{0.865\textwidth}
         \centering
        \includegraphics[width=1\linewidth]{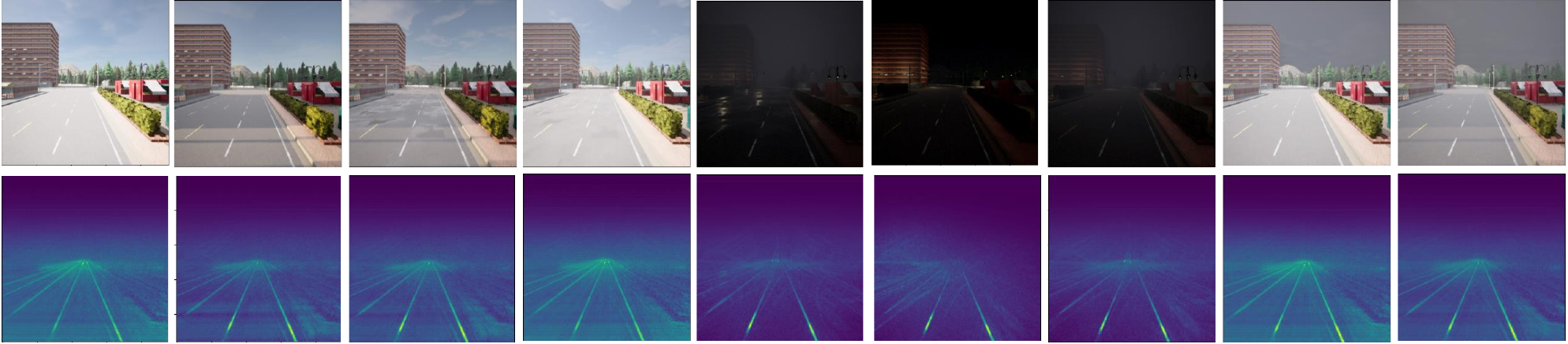}
         \caption{The system outputs for lane detection in 'Map02' under different environmental conditions.}
         \label{fig:gen_map02}
     \end{subfigure}
     \hfill
    \begin{subfigure}[b]{0.865\textwidth}
         \centering
        \includegraphics[width=1\linewidth]{ 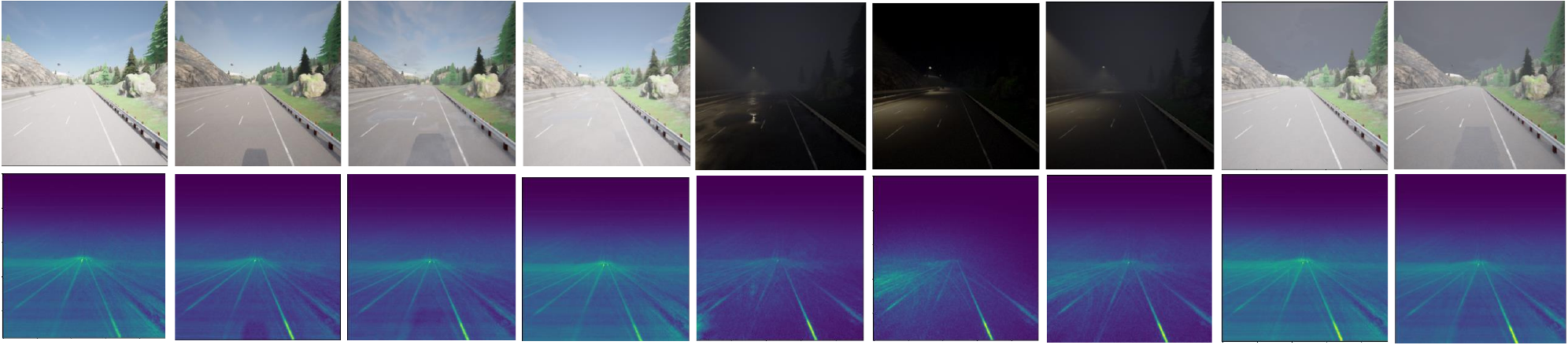}
         \caption{The system outputs for lane detection in 'Map03' under different environmental conditions.}
         \label{fig:gen_map03}
     \end{subfigure}
     \hfill
    \begin{subfigure}[b]{0.865\textwidth}
         \centering
        \includegraphics[width=1\linewidth]{ 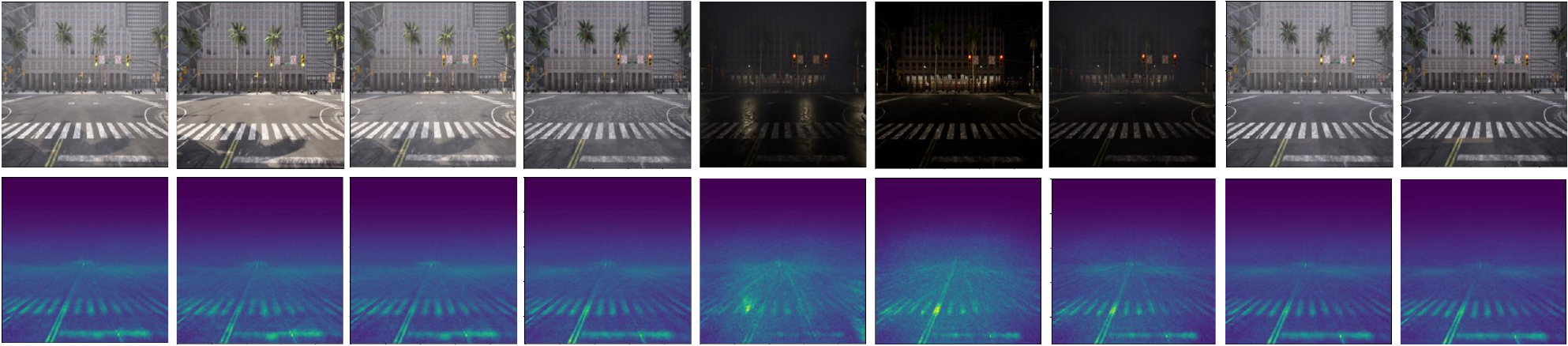}
         \caption{The system outputs for lane detection in 'Map04' under different environmental conditions.}
         \label{fig:gen_map04}
     \end{subfigure}
     
    \caption{The generalizability test of the proposed model with different maps under different environmental conditions. }
    \vspace{-4mm}
    \label{fig:gen_test}
\end{figure*}

\subsubsection{\textbf{Evaluations on Simulated Scenes in CARLA}}
\label{sec:lane_carla}

\begin{figure}
     \centering
     \includegraphics[width=0.45\textwidth]{ 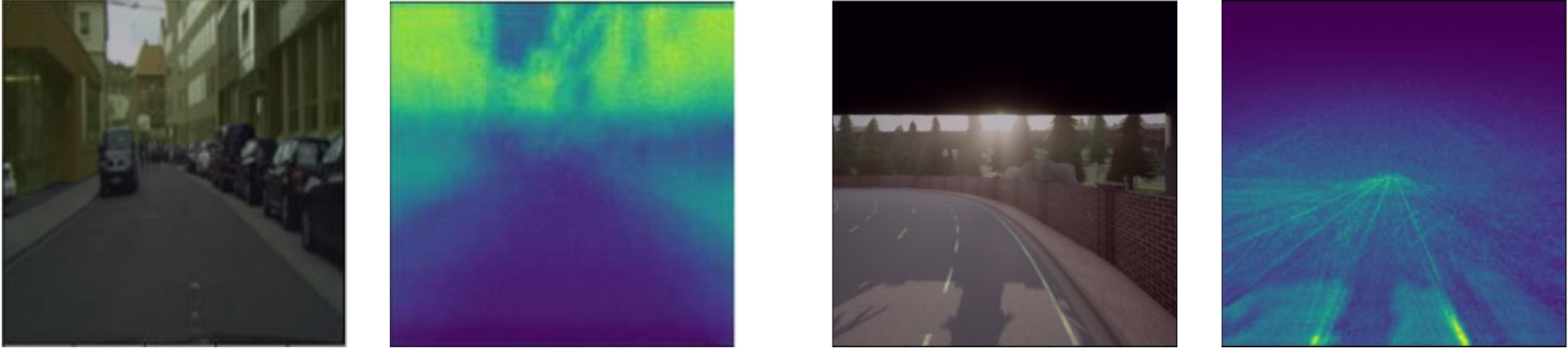}
        \caption{{DONN model is highly sensitive to light distribution. Light reflection and shadow can cause prediction noise.}}
        \label{fig:discussion_samples}
        \vspace{-6mm}
\end{figure}
% for the prepared dataset, we have 2700 for training, for each map with each weather condition, we have 100 for inference

% python3 auto_sim.py --depth 8 --sys-size 480 --model-save-path ./model_sim_0317/ --image-save-path ./vis_sim_0317/ --epochs 802 --lr 0.15 --datapath ./carla_sim_0317/ --batch-size 64

In this section, we further evaluate the generalizability of our RGB-channel DONN system with more complex dataset under diverse environmental conditions. 
%We record the simulated scenes along driving on the road in different maps under different weather conditions at different times of the day (as shown in Figure \ref{fig:gen_test}) from the open-sourced driving simulator CARLA. 
The model is implemented with the system size of $480 \times 480$. Due to the extreme imbalance between the object and background, we implement the weighted binary cross-entropy loss to train the model as described in \cite{lee2023end}.  
%\begin{equation}\label{equ:bce_loss}
%\small
%\begin{aligned}
%\mathcal{L}_{\text{weighted\_BCE}} = 
%& - \frac{P}{N + P} \sum_{i=1}^{N} \sum_{j=1}^{N} I_{\text{GT}}^{ij} \log(I_{\text{det}}^{ij}) \\
%& - \frac{N}{N + P} \sum_{i=1}^{N} \sum_{j=1}^{N} (1 - I_{\text{GT}}^{ij}) \log(1 - I_{\text{det}}^{ij})
%\end{aligned}
%\end{equation}
We first introduce the customized dataset generation by CARLA in Section \ref{sec:carla_data_intro}, and we present the generalizability of the trained DONN model in Section \ref{sec:carla_gen} to different maps, weather conditions, and times of the day.

\paragraph{\textbf{Dataset Generation}}\label{sec:carla_data_intro}

In this section, we introduce the details of the customized simulated driving dataset generated from simulations in CARLA. As shown in Figure \ref{fig:gen_test}, we collect samples from different maps, where 'Map01' and 'Map02' are used for training, while 'Map03' and 'Map04' are used for evaluations only, under different weather conditions, including 'Clear', 'Rainy', and 'Cloudy', at different times of the day, including 'Noon', 'Sunset', and 'Night'. %The model is trained with the dataset including simulations with weather conditions 'Clear' and 'Rainy', and time of the day 'Noon'. The weather condition 'Cloudy' and the time of the day 'Night' are used for evaluations only. 

For training dataset, we collect $800$ samples for each map under each environment condition including 'Clear' and 'Rainy' at 'Noon' and 'Sunset', 'DustStorm', and 'Default' weather conditions in the simulator. Thus, the total number of samples in the training dataset is $6 \times 2 \times 800 = 9,600$. For the evaluation dataset, we collect $200$ samples for each environment condition for all four maps, to evaluate the model's generalizability. Examples are shown in Figure \ref{fig:gen_test}.  

\paragraph{\textbf{Generalizability Evaluation}} \label{sec:carla_gen}
%In this section, we present our evaluation results of the trained DONN model to demonstrate its generalizability to different environment conditions including different maps, different weather conditions, and different times of the day. The first four columns in Figure \ref{fig:gen_map01} and Figure \ref{fig:gen_map02} are new evaluations samples with the model trained with the same maps and environmental conditions. Our model can detect the lane marking successfully while the model is very sensitive to the light distribution. For example, in 'Clear Noon' in Figure \ref{fig:gen_map01}, the model detects the shade of the tree while in 'Rainy Sunset', it detects the reflection of the water on the road. 

This section presents the evaluation of the trained DONN model to assess its generalizability across varying environments, including different maps, weather conditions, and times of day. The first four columns in Figures \ref{fig:gen_map01} and \ref{fig:gen_map02} show unseen samples with the same training maps and conditions. The model effectively detects lane markings, while showing high sensitivity to light distribution. For example, under 'Clear Noon', it identifies tree shadows, while, under 'Rainy Sunset,' it detects water reflections on the road, resulting in noise for lane detection.

%\noindent\textbf{Generalizability to different maps} -- As shown in the first four columns of 'Map03' and 'Map04' in Figure \ref{fig:gen_map03} and Figure \ref{fig:gen_map04}, the model trained with 'Map01' and 'Map02' can detect the lane markings in new maps successfully.

\noindent\textbf{Generalizability to Different Maps} -- As shown in the first four columns of 'Map03' and 'Map04' in Figures \ref{fig:gen_map03} and \ref{fig:gen_map04}, the model trained on 'Map01' and 'Map02' successfully detects lane markings in unseen maps.

%\noindent\textbf{Generalizability to different weather conditions} -- As shown in the columns 'Cloudy Noon' and 'Cloudy Sunset' in Figure \ref{fig:gen_test}, the model trained with 'Clear' and 'Rainy' weather conditions can detect the lane marking successfully with 'Cloudy'. The weather 'Clear' indicates the most direct light distribution regarding the time of the day. For example, for 'Clear Noon', the average light intensity of the whole image will be high and there will be a clear shade of the tree on the road. For 'Clear Sunset', it will be a spot light source, causing uneven light distribution. The weather 'Rainy' indicates a reflection of water on the road. The weather 'Cloudy' indicates relatively low average light intensity and less light distribution imbalance, e.g., for the strong spot light source in 'Sunset'. Compared with 'Clear' and 'Rainy', 'Cloudy' samples result in predictions with more even light distribution while the lane markings can be differentiated from the background.  

\noindent\textbf{Generalizability to different weather conditions} -- As shown in Figure \ref{fig:gen_test}, 'Clear' conditions exhibit direct illumination from the light source, with minimal diffuse reflection or attenuation in brightness. 'Rainy' conditions introduce water reflections on the road. 'Cloudy' conditions yield lower overall brightness and reduced lighting imbalance, particularly at sunset. As shown in the 'Cloudy Noon' and 'Cloudy Sunset' columns, the model trained on 'Clear' and 'Rainy' conditions successfully detects lane markings under 'Cloudy' weather. %Compared to 'Clear' and 'Rainy,' 'Cloudy' conditions result in more uniform lighting while maintaining lane marking visibility.

%\noindent\textbf{Generalizability to different time of the day} -- As shown in columns with 'Night' in Figure \ref{fig:gen_test}, the model trained with 'Noon' and 'Sunset' can detect the lane markings in 'Night' successfully. The time 'Noon' indicates high average light intensity over the whole image, which can result in clear shades and ambiguous boundary of the land markings on the road. The time 'Sunset' indicates a spot light source in the images, resulting in uneven light distributions. The time 'Night' indicates low light intensity over the image while the streetlight in the night can result in extremely uneven light distribution. For example, in 'Rainy Night' shown in Figure \ref{fig:gen_map03} and Figure \ref{fig:gen_map04}, the water on the road reflects the streetlight and results in the noise for model predictions.  

\noindent\textbf{Generalizability to different time of the day} -- As shown in the 'Night' columns in Figure \ref{fig:gen_test}, the model trained on 'Noon' and 'Sunset' successfully detects lane markings at 'Night'. 'Noon' conditions feature high overall brightness, leading to clear shadows and potentially ambiguous lane boundaries. 'Sunset' introduces a localized light source, causing uneven illumination. 'Night' conditions exhibit low overall brightness, with streetlights creating extreme lighting variations. For example, in 'Rainy Night' (Figures \ref{fig:gen_map03} and \ref{fig:gen_map04}), water reflections from streetlights introduce noise in model predictions.

\section{Discussions}
\label{sec:discussion}

We propose a novel free-space all-optical DONN system for image processing in machine learning in this work. The general observations of the DONN model for image segmentation from our work indicate that the DONN model shows: 
\textbf{(1) High sensitivity to light distribution.} The DONN model's performance is highly influenced by the \textbf{uniformity of environmental lighting}. A more uniform light distribution generally leads to improved image processing performance as demonstrated from the comparisons between Section \ref{sec:lane_toy} and Section \ref{sec:carla_gen}, where the DONN model exhibits reduced noise and improved lane detection performance with the indoor track dataset. %For instance, as shown in Figure \ref{fig:city_seg_rgb1}, when the light distribution is even—even if the image appears dark—the model produces more detailed and accurate segmentation results. Additionally, performance under 'Sunset' conditions, where a strong spot light (the sun) is present, is inferior to conditions with more uniform lighting or weaker spot lights, such as 'Night' conditions.
Moreover, the \textbf{high-contrast lighting distribution} such as water reflection and shadows, can confuse the model. As illustrated in Figures \ref{fig:discussion_samples}, reflections on building glass surfaces and the shadow on the road can result in noise for image segmentation. %Since the DONN system operates purely by manipulating light signals, with input information encoded via light intensity, it tends to suffer from weak feature compression capabilities, and overfit to the details with significant light contrast.
\textbf{(2) Necessity for improved post-processing optical binarization methods.} The diffraction process can cause uneven light distribution over the whole image. Although the system output preserves segmentation boundaries, coarse binarization techniques can eliminate delicate segmentation details. \textbf{Developing a learnable model for adaptive post-processing tailored to varying outputs is critical to optical image segmentation.} {Moreover, \textbf{(3) the hardware realization of DONN systems is critical for practical free-space all-optical computing}, which requires advanced design and fabrication technologies such as on-chip integration and optoelectronic computing~\cite{luo2022metasurface,zhou2021large}.}

%%%%%%%%%%%%%%%%%%%%%%%%%%%%%%%%%%%%%%%%%%%%
\section{Conclusion}
\label{sec:conclusion}

In this work, we propose a novel DONN architecture for RGB image processing, targeting tasks such as image segmentation and lane detection in autonomous driving systems. The model processes red ("R"), green ("G"), and blue ("B") components of a RGB image separately with all-optical computing channels, reducing ADC-related computational overhead and energy consumption by processing images in the optical domain.
Furthermore, the DONN model offers high system throughput and ultrafast computation at the speed of light, making it a strong candidate for perception models in autonomous driving systems with power constraints for hardware deployments.
We evaluate the model with image segmentation task using the Cityscapes dataset and lane detection task using customized indoor track dataset and CARLA-simulated driving dataset. We further evaluate the model's generalizability diverse environments, including different maps, weather conditions, and times of day with dataset from CARLA simulations, demonstrating its potential for real-world deployment in autonomous driving scenarios.

\newpage
\bibliographystyle{acm}
\bibliography{ref}

\end{document}